\documentclass{article}

\PassOptionsToPackage{numbers}{natbib}
\usepackage[preprint]{neurips_2026}

\usepackage[utf8]{inputenc}
\usepackage[T1]{fontenc}
\usepackage{hyperref}
\usepackage{url}
\usepackage{booktabs}
\usepackage{amsfonts}
\usepackage{amssymb}
\usepackage{float}
\usepackage{nicefrac}
\usepackage{microtype}
\usepackage{xcolor}
\usepackage{graphicx}

\title{Spectra: A Rules-Driven LLM Pipeline for Automated KYC Document Processing}

\author{
  Miray Wahib \quad
  Ethan Tran \quad
  Rea Mourad \quad
  Mira Muti \quad
  Nikita Dvornik \\[0.5em]
  Royal Bank of Canada\thanks{This work was completed as part of the RBC Amplify program.} \\[0.3em]
  \texttt{miray.wahib@mail.mcgill.ca} \quad
  \texttt{ethan.tran@mail.mcgill.ca} \quad
  \texttt{mourad.rea@gmail.com} \\
  \texttt{mira.muti@mail.mcgill.ca} \quad
  \texttt{nikita.dvornik@rbc.com}
}

\begin{document}

\raggedbottom

\maketitle

\begin{abstract}
Know Your Client (KYC) onboarding in capital markets requires analysts to manually classify documents, extract structured data from heterogeneous sources, and validate compliance against complex regulatory policies. This process requires significant analyst time per client, with end-to-end onboarding often stretching to multiple weeks due to sequential handoffs. In this work, we analyze an on-boarding process and find that it comprises repeatable components well-suited to AI automation. We therefore propose a restructured workflow to be amenable to automation: we consolidate the traditional four-party process into two parties that share most of the work and can be automated together, eliminating intermediate handoffs that compound delays. To automate the remaining steps, we introduce Spectra, an AI-assisted document processing platform that combines a structured rules engine with LLM-based classification, extraction, and validation agents. The rules engine encodes compliance policy as a queryable database, enabling focused context injection that reduces token usage while improving extraction precision. Rather than a single monolithic prompt, the system decomposes document processing into isolated, auditable stages—each optimized independently and traceable to specific policy clauses. In evaluation on real KYC documents, Spectra achieves 100\% classification accuracy and 89.4\% extraction accuracy. Human review burden dropped by 96\%.
\end{abstract}

\section{Introduction}

Client onboarding in capital markets is a compliance-critical process governed by Know Your Client (KYC) and Anti-Money Laundering (AML) regulations. Before any institutional client can trade with a bank, they must provide documentation proving their legal identity, beneficial ownership structure, regulatory status, and business activities, among other requirements that vary by entity type and jurisdiction. The risk of inadequate KYC is substantial: systemic compliance failures can result in regulatory penalties, reputational damage, and in severe cases, facilitation of financial crime.

The core bottlenecks are threefold. First, 20+ documents per case must be classified against a taxonomy of 50+ document types, each with jurisdiction-specific requirements. Second, field-by-field data extraction from PDFs---entity names, registration numbers, addresses, authorized signatories---is tedious and error-prone. Third, validation against policy requires cross-referencing 400-page compliance documents to determine which rules apply to a given entity type, jurisdiction, and product combination.

Recent advances in large language models (LLMs) have fundamentally changed what is possible for document understanding tasks. Multi-modal LLMs can now parse complex document layouts, extract structured data from heterogeneous formats, and follow nuanced instructions that would have required brittle rule-based systems or extensive template engineering just a few years ago. Crucially, these models can handle the variability inherent in KYC documents—certificates of incorporation from different jurisdictions, regulatory filings with inconsistent formatting, corporate resolutions with varying structures—without requiring per-template customization. However, applying a single monolithic LLM call to the full KYC workflow is insufficient: context windows cannot accommodate hundreds of pages of policy, errors are difficult to localize, and regulatory audit trails require traceability to specific policy clauses. This motivates an agentic architecture where specialized agents handle classification, extraction, and validation as isolated stages—each with focused context, structured outputs, and independent optimization paths. This approach builds on recent work demonstrating that LLMs perform more reliably when reasoning and acting are interleaved in structured loops \cite{yao2023react}, when models can invoke external tools to ground their outputs \cite{schick2023toolformer}, and when verbal feedback enables iterative refinement \cite{shinn2023reflexion}.

We present Spectra, an AI-assisted KYC workflow platform that addresses these bottlenecks through three technical contributions:

\begin{enumerate}
    \item A \textbf{structured rules engine} that encodes compliance policy as a queryable database, enabling context-aware schema generation and deterministic checklist creation without LLM calls.
    \item A \textbf{multi-agent document processing pipeline} that decomposes classification, extraction, and validation into isolated stages with structured outputs and explicit confidence scoring.
    \item An \textbf{evaluation framework} demonstrating 100\% classification accuracy on real KYC documents and 89.4\% extraction accuracy on synthetic test documents, with per-field and per-document-type breakdowns that identify remaining gaps.
\end{enumerate}

The system presents a workflow with two teams: a Gathering Team that collects and processes documents, and an Approval Team that reviews flagged items. By surfacing extractions below 100\% confidence for human review, Spectra aims to reduce analyst burden while preserving the auditability required for regulatory compliance.

\section{Approach}

This section presents our approach: encoding compliance policy in a queryable rules engine and decomposing document processing into a multi-agent pipeline.

\subsection{AI-Native KYC Workflow}

\begin{figure}[H]
  \centering
  \includegraphics[width=0.9\linewidth]{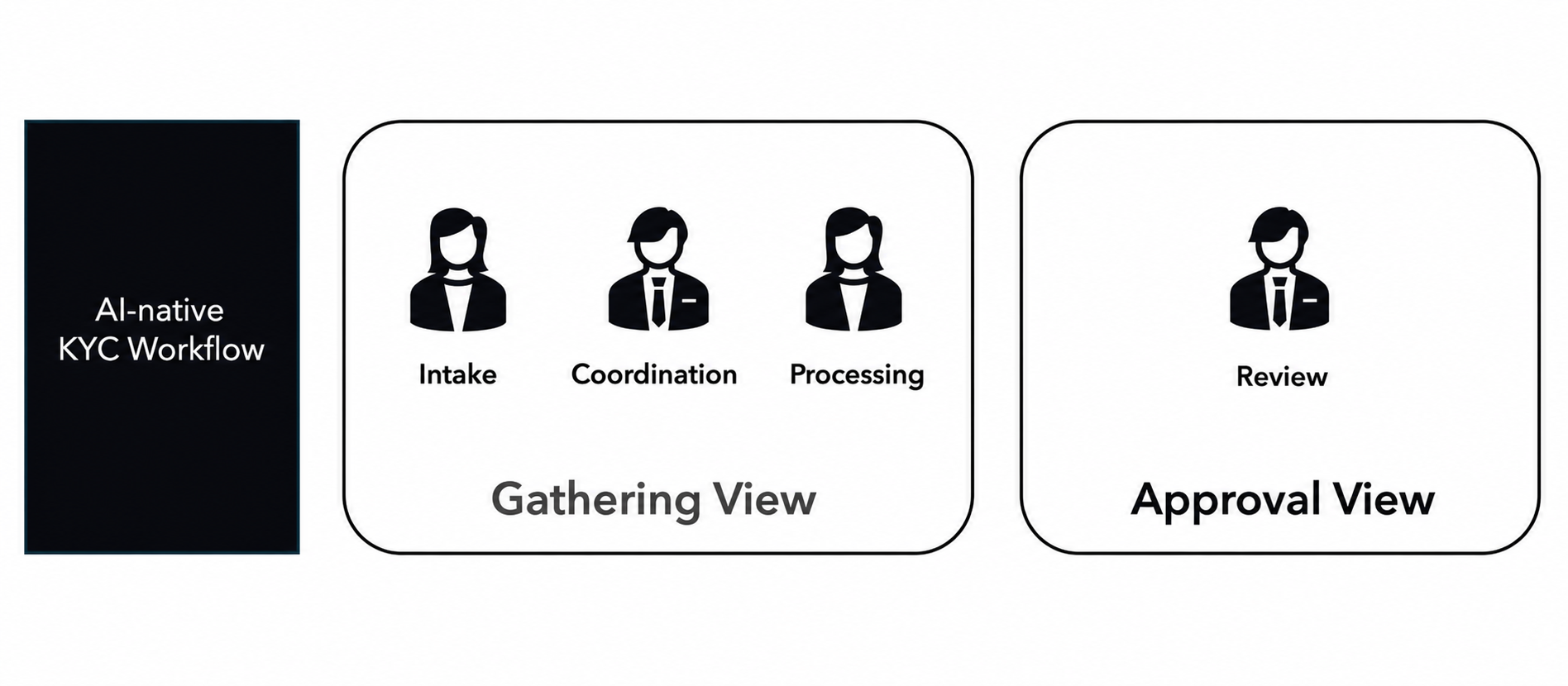}
  \caption{AI-native KYC workflow. Gathering View presents Intake, Coordination, and Processing in a shared workspace for document collection and AI-assisted processing. Approval View presents Review for exception handling and final sign-off.}
  \label{fig:workflow}
\end{figure}

Spectra presents an AI-native KYC workflow with two team views (Figure~\ref{fig:workflow}). The \textbf{Gathering View} brings Intake, Coordination, and Processing into a shared workspace. Within this view, users search for or create client records, generate the document checklist from the rules engine, upload documents, and inspect outputs from the classification, extraction, and validation agents before submitting the case for approval.

The \textbf{Approval View} supports Review. It surfaces data points below 100\% confidence and manual overrides alongside source evidence and policy references. Reviewers can approve the case or return it to the Gathering View with specific deficiencies.

The journey through Spectra follows a defined lifecycle. A case begins in \texttt{draft} status during initial setup, moves to \texttt{pending\_approval} when the Gathering Team submits it, enters \texttt{in\_review} when an Approval Team member claims it, and resolves to either \texttt{approved} (client can begin trading) or \texttt{returned} (sent back to Gathering Team with specific deficiencies noted). This lifecycle ensures clear ownership at each stage and provides the audit trail required for regulatory compliance.

\subsection{Problem Formulation}

The AI automation within this workflow can be decomposed into three sequential tasks:

\textbf{Classification.} Given an uploaded document $d$ and a journey-specific candidate set $C = \{c_1, c_2, ..., c_k\}$ of acceptable document types, predict the document type $\hat{c} \in C$ with associated confidence $p(\hat{c}|d)$.

\textbf{Extraction.} Given a classified document $(d, \hat{c})$ and a dynamically generated extraction schema $S_{\hat{c}}$ specifying required fields, extract structured values $\{(f_i, v_i, p_i)\}$ where $f_i$ is the field name, $v_i$ is the extracted value, and $p_i$ is the extraction confidence.

\textbf{Validation.} Given extracted data points and applicable policy rules $R$, determine whether each value passes, fails, or triggers an exemption that modifies the journey checklist.

A naive approach would concatenate all policy documents into a single prompt and ask an LLM to perform all three tasks simultaneously. This fails for several reasons: (1) policy documents exceed context limits, (2) errors cannot be localized to a specific stage, (3) retry logic must restart the entire pipeline, and (4) validation reasoning lacks traceability to specific policy clauses.

\subsection{Rules Engine Architecture}

The foundation of Spectra is a structured rules engine implemented as a PostgreSQL database with the following core tables:

\begin{itemize}
    \item \texttt{documents}: Document type definitions including codes, names, descriptions, and invalidation rules.
    \item \texttt{data\_requirements}: Field definitions with JSON schemas specifying expected types and formats.
    \item \texttt{document\_requirements}: Requirement definitions by jurisdiction, entity type, and product, linking to acceptable document types.
    \item \texttt{exemptions}: Conditions that add or remove requirements based on extracted data.
\end{itemize}

Join tables (\texttt{must\_contain}, \texttt{acceptable\_documents}, \texttt{doc\_data\_requirements}) encode the relationships between documents, fields, and requirements. The complete schema is detailed in Appendix~\ref{appendix:rulesengine}.

The rules engine serves two purposes. First, it enables \textbf{deterministic checklist generation}: given a client's jurisdiction, entity type, and product, the system queries the database to produce the exact set of required documents without any LLM calls. Second, it provides \textbf{focused context injection}: when an LLM agent processes a document, it receives only the relevant document definitions, required fields, and applicable validation rules—not the entire set of multi-page documents uploaded to the system.

\subsection{Multi-Agent Pipeline}

\begin{figure}[H]
  \centering
  \includegraphics[width=\linewidth]{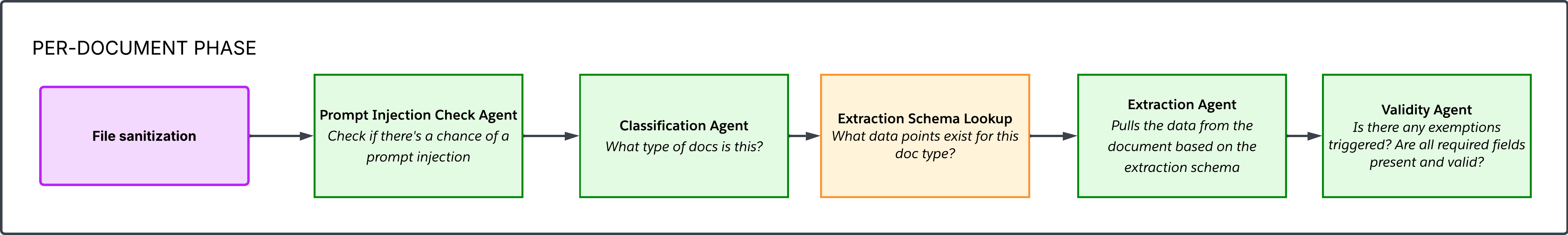}
  \caption{Per-document processing pipeline. Each uploaded document passes through six stages: file sanitization, injection detection, classification, schema lookup, extraction, and validation. Green stages involve LLM agents; the orange stage (schema lookup) is a deterministic database query against the Rules Engine. This decomposition enables independent optimization and auditability at each stage.}
  \label{fig:pipeline}
\end{figure}

The document processing pipeline consists of six sequential stages, each implemented as an isolated service with structured inputs and outputs. Figure~\ref{fig:pipeline} details the per-document processing flow.

\textbf{Stage 1: File Sanitization.} Before any AI processing, uploaded files undergo validation: MIME type detection (not just extension checking), size limits (50 MB max), filename sanitization, and text extraction via PyMuPDF or Tesseract OCR. The extracted text is stored as an S3 sidecar for downstream reuse.

\textbf{Stage 2: Injection Detection.} Documents from external sources may contain adversarial instructions designed to manipulate LLM behavior. Spectra applies two detection layers: regex pattern matching for known attack phrases (e.g., "ignore all instructions") and a semantic classifier for subtler attacks. Flagged documents continue through the pipeline but are surfaced for security review.

\textbf{Stage 3: Classification.} The classifier receives the extracted text and a filtered candidate set derived from the journey checklist. The LLM outputs a predicted document code, confidence score, and reasoning.

\textbf{Stage 4: Schema Lookup.} Once the document type is known, the rules engine generates the extraction schema as a union of: (1) \texttt{must\_contain} fields for the document type, and (2) \texttt{doc\_data\_requirements} fields for the matched requirement. This schema is a JSON object specifying field names, types, and descriptions. Notably, this stage requires no LLM call—it is a deterministic database query.

\textbf{Stage 5: Extraction.} The extraction agent receives the document text and generated schema. It returns structured values with per-field confidence and source evidence (page number, bounding box, text snippet).

\textbf{Stage 6: Validation.} Each extracted data point is validated against policy rules. Validation checks include: null detection, completeness assessment, validity against policy, and exemption evaluation. The output includes a pass/warning/fail status, policy reference citation, and reasoning.

This decomposition enables independent optimization of each stage. When extraction accuracy is low for a specific document type, we can adjust prompts or schemas for that type without affecting classification. When validation reasoning is unclear, we can improve the policy encoding without retraining extractors.

The pipeline leverages two LLM backends. Document classification and extraction (Stages 3 and 5) use LlamaCloud \cite{llamaindex2024}, a managed document processing service by LlamaIndex that provides specialized endpoints for document understanding tasks. Validation and injection detection (Stages 2 and 6) use OpenAI's o3-mini \cite{openai2025o3mini} and GPT-4.1-mini \cite{openai2024gpt41} models, accessed through the internal LLM gateway.

\subsection{Human-in-the-Loop Design}

Spectra does not aim to remove humans from KYC approval. Final approval authority remains with the Approval Team, as required by banking regulations (four-eye check principle). Instead, the system reduces review burden by:

\begin{itemize}
    \item Auto-approving data points with 100\% confidence (no review needed).
    \item Flagging any data point below 100\% confidence for human review.
    \item Providing clickable source evidence so reviewers can verify extractions directly against the document.
    \item Recording override justifications when analysts disagree with system outputs.
\end{itemize}

In our controlled test case, this reduced required review items from 206 to 8—a 96\% reduction in manual verification workload.

\subsection{Scalability Considerations}

Spectra's architecture separates policy encoding from AI execution, enabling a plug-and-play model for enterprise-wide deployment.

\textbf{Policy complexity.} The rules engine schema accommodates jurisdiction-specific requirements, entity-type variations, and product-based exceptions without modifying the AI pipeline. Adding a new regulatory requirement involves database inserts, not model retraining or prompt engineering. The current implementation encodes requirements across multiple jurisdictions and entity types; extending coverage requires only schema population.

\section{Evaluation}

In this section, we evaluate Spectra across two dimensions: accuracy and efficiency. For accuracy, we measure the performance of classification (Stage 3) and extraction (Stage 5) using standard metrics including precision, recall, F1 score, and exact match rate. We further disaggregate extraction performance by document type and field type to identify specific areas requiring improvement. For efficiency, we report per-document processing time and estimate end-to-end time savings based on the workflow transformation described in Section 2.1.

\subsection{Dataset}

We evaluated Spectra's classification capability on a corpus of 21 real KYC documents spanning 9 document types:
\begin{itemize}
    \item Corporate resolutions (Type A)
    \item Articles of incorporation (Type B)
    \item Regulatory filings (Type C)
    \item Proof of regulation (Type D)
    \item AML policy letters (Type E)
    \item AML questionnaires (Type F)
    \item Sanctions letters (Type G)
    \item Revenue declarations (Type H)
    \item Name change certificates (Type I)
\end{itemize}

For extraction evaluation, we created 15 synthetic test documents with 141 manually annotated field values across 27 field types, including entity names, dates, addresses (with subfields), registration numbers, and structured nested objects (e.g., authorized signatories with name, title, and authorization scope).

\subsection{Classification Results}

The classification agent achieved perfect accuracy on the test set:

\begin{table}[h]
  \caption{Document Classification Performance}
  \label{tab:classification}
  \centering
  \begin{tabular}{lcccc}
    \toprule
    Metric & Value \\
    \midrule
    Total Documents & 21 \\
    Correctly Classified & 21 \\
    Accuracy & 100\% \\
    Macro Precision & 1.00 \\
    Macro Recall & 1.00 \\
    Macro F1 & 1.00 \\
    Average Confidence & 99.3\% \\
    Average Time & 11.6 seconds \\
    \bottomrule
  \end{tabular}
\end{table}

All 9 document types achieved perfect precision and recall. 81\% of documents were classified with 100\% confidence; the remaining 19\% were at 95\% confidence.

While these results are strong, we note limitations: (1) the test set is small (n=21), (2) all documents were relatively clean PDFs without significant OCR challenges, and (3) the candidate set filtering may reduce the effective classification difficulty. Future work should evaluate on a larger, more diverse corpus including scanned documents with OCR artifacts.

\subsection{Extraction Results}

Extraction performance was evaluated on 15 documents with 141 annotated fields:

\begin{table}[h]
  \caption{Data Extraction Performance}
  \label{tab:extraction}
  \centering
  \small
  \begin{tabular}{lclc}
    \toprule
    \textbf{Metric} & \textbf{Value} & \textbf{Metric} & \textbf{Value} \\
    \midrule
    Accuracy & 89.4\% & Exact Match Rate & 89.4\% \\
    Precision & 86.6\% & Partial Match Rate & 94.3\% \\
    Recall & 94.7\% & Avg. Confidence & 43.3\% \\
    F1 Score & 90.5\% & Avg. Time & 11.0s \\
    \bottomrule
  \end{tabular}
\end{table}

The confusion matrix breaks down as: 71 true positives, 55 true negatives, 11 false positives, and 4 false negatives.

\begin{table}[h]
  \caption{Performance by Document Type and Field}
  \label{tab:performance_breakdown}
  \centering
  \small
  \begin{tabular}{llccc}
    \toprule
    \textbf{Category} & \textbf{Name} & \textbf{Acc.} & \textbf{F1} & \textbf{N} \\
    \midrule
    \multicolumn{5}{l}{\textit{By Document Type}} \\
    & Type B (Articles of Incorp.) & 100\% & 1.00 & 30 \\
    & Type E (AML Policy) & 100\% & 1.00 & 27 \\
    & Type J (Business Description) & 95.6\% & 0.97 & 45 \\
    & Type C (Regulatory Filing) & 69.7\% & 0.55 & 33 \\
    & Type A (Corporate Resolution) & 50.0\% & 0.67 & 6 \\
    \midrule
    \multicolumn{5}{l}{\textit{By Field (Best)}} \\
    & \texttt{entity\_legal\_name} & 100\% & --- & 12 \\
    & \texttt{country\_of\_incorporation} & 100\% & --- & 12 \\
    & \texttt{incorporation\_number} & 100\% & --- & 12 \\
    & \texttt{registered\_address} & 100\% & --- & 3 \\
    & \texttt{business\_description} & 100\% & --- & 3 \\
    \midrule
    \multicolumn{5}{l}{\textit{By Field (Worst)}} \\
    & \texttt{powers\_to\_bind} & 0\% & --- & 3 \\
    & \texttt{products\_product\_type} & 33\% & --- & 3 \\
    & \texttt{naics\_code} & 67\% & --- & 3 \\
    & \texttt{formation\_date} & 75\%* & --- & 4 \\
    \bottomrule
    \multicolumn{5}{l}{\footnotesize *75\% exact match, 100\% partial—formatting differences suggest a normalization layer could recover these.} \\
  \end{tabular}
\end{table}

Type A (corporate resolutions) had the worst performance at 50\% accuracy. Analysis revealed that the \texttt{powers\_to\_bind} field consistently failed: the model extracted structured data when the ground truth was null, suggesting the schema definition or prompt needs refinement for this edge case.

\subsection{Failure Analysis}

We identified three primary failure modes:

\textbf{Schema mismatch for complex fields.} The \texttt{powers\_to\_bind} field has a complex nested structure that the current prompt instructions do not adequately constrain. Hard schema enforcement does not yet exist; the model inferred an incorrect structure when the field should be null, drawing values from related fields elsewhere in the document.

\textbf{Vocabulary ambiguity.} The \texttt{products\_product\_type} field lacks a controlled vocabulary in the schema. The model extracts descriptive text (e.g., "Private Equity") when the expected value uses a different taxonomy.

\textbf{Cross-field inference.} Some fields (e.g., \texttt{regulated\_entity\_information}) require synthesizing information across multiple document sections. The model sometimes over-extracts, filling subfields that are present in the document but not requested for the specific requirement.

\begin{figure}[H]
  \centering
  \includegraphics[width=\linewidth]{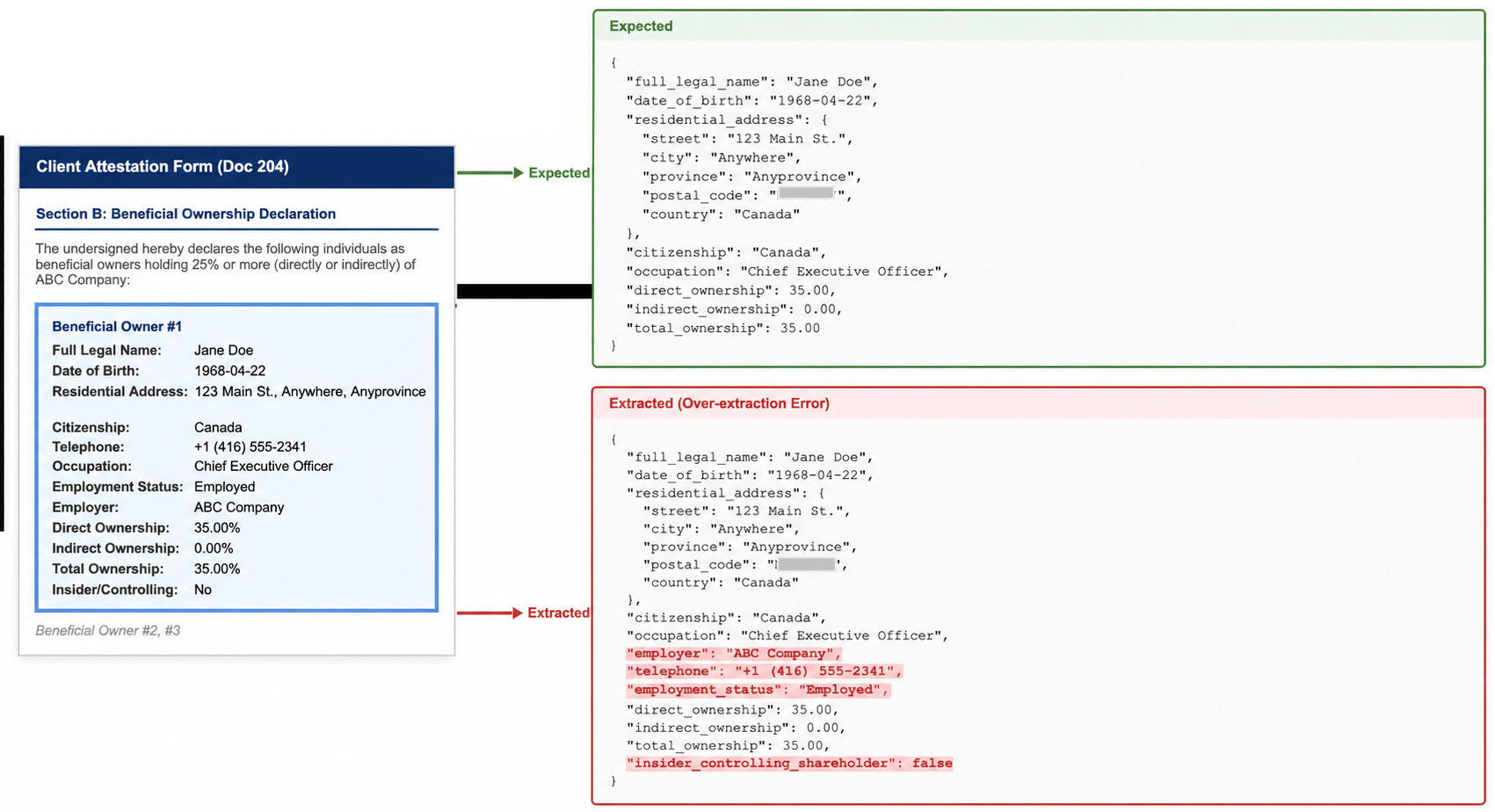}
  \caption{Extraction comparison for beneficial owner data from a synthetic Client Attestation Form. The highlighted source section (left) contains both schema-required fields and additional document fields. The expected extraction (top right) includes only fields specified in the extraction schema. The actual extraction (bottom right) demonstrates an over-extraction error: the model included fields present in the document (highlighted in red) that were not specified in the schema.}
  \label{fig:examples}
\end{figure}

Figure~\ref{fig:examples} illustrates this over-extraction pattern on a Client Attestation Form. The model correctly extracted all schema-required fields (name, date of birth, address, citizenship, occupation, ownership percentages) but also included adjacent fields from the document (employer, telephone, employment status, insider status) that were not part of the extraction schema. This suggests the failure is not due to document quality or parsing issues, but rather to insufficient schema constraints that would signal which fields to extract versus which to ignore despite their proximity in the source document.

\section{Conclusion}

We presented Spectra, an AI-assisted KYC document processing system that combines a structured rules engine with LLM-based classification, extraction, and validation agents. By encoding compliance policy in a queryable database and decomposing document processing into isolated, auditable stages, Spectra achieves high accuracy while maintaining full auditability.

Challenges remain in handling certain complex document types and fields, and the system requires further evaluation on larger, more diverse document corpora. However, the architecture is designed for iterative improvement: the rules engine can be extended without retraining models, and individual pipeline stages can be optimized independently.

\section*{Acknowledgments}

We thank the domain experts who provided guidance throughout this work.

\appendix

\section{Rules Engine Schema}
\label{appendix:rulesengine}

\begin{figure}[H]
  \centering
  \includegraphics[width=\linewidth]{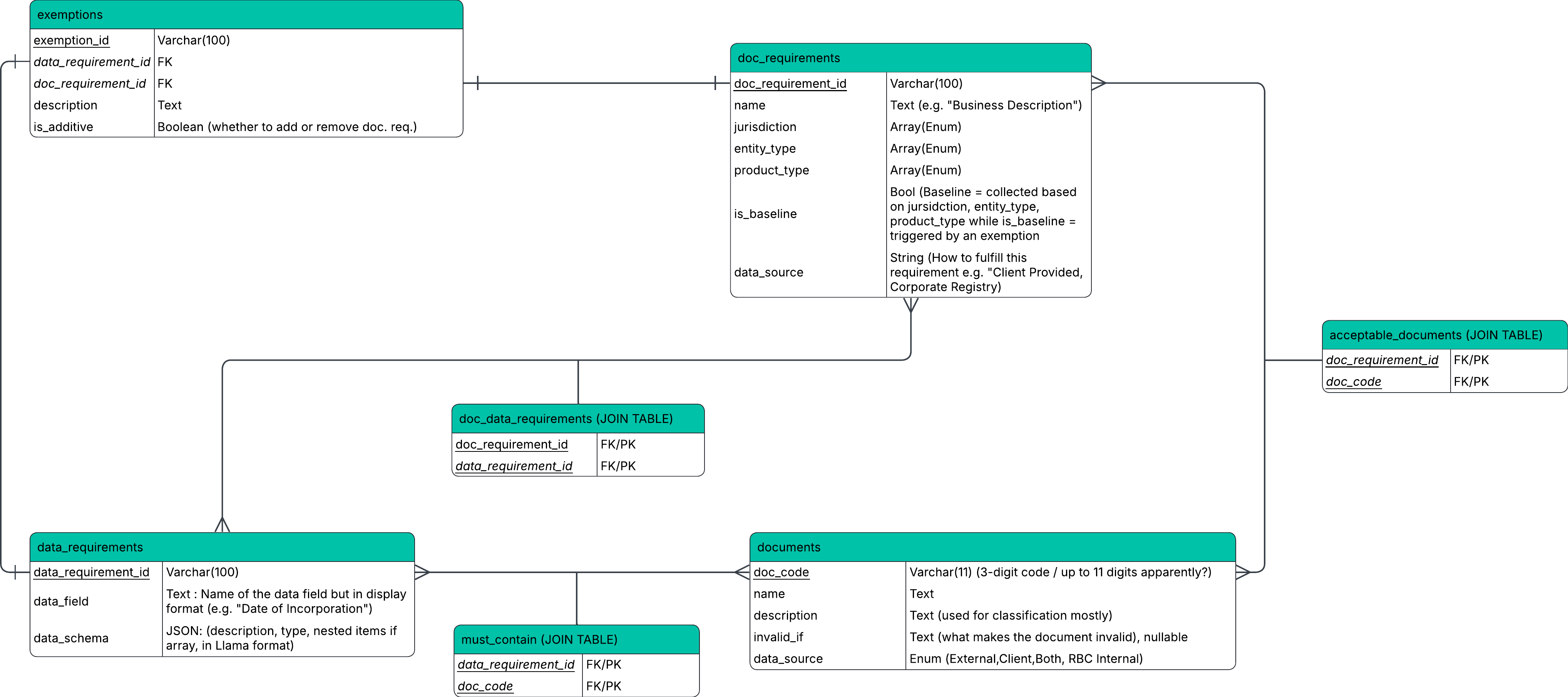}
  \caption{Rules engine entity-relationship diagram. Core tables (teal) define documents, requirements, data fields, and exemptions. Join tables (white) encode many-to-many relationships that determine which documents satisfy which requirements and which fields must be extracted from each document type.}
  \label{fig:rulesengine}
\end{figure}

Figure~\ref{fig:rulesengine} presents the complete entity-relationship diagram for Spectra's rules engine. This schema encodes KYC compliance policy in a structured, queryable format—eliminating the need to pass hundreds of pages of policy documents to LLM agents.

\subsection{Core Tables}

\textbf{\texttt{doc\_requirements}.} This is the central table that defines what documentation is required for a given onboarding scenario. Each row represents a requirement (e.g., "Business Description") scoped by three dimensions: \texttt{jurisdiction} (array of applicable regions), \texttt{entity\_type} (array of entity classifications), and \texttt{product\_type} (array of products being onboarded). The \texttt{is\_baseline} flag distinguishes standard requirements (collected for all matching clients) from conditional requirements triggered by exemptions. The \texttt{data\_source} field specifies how the requirement should be fulfilled—whether the client must provide documentation, whether it can be retrieved from a corporate registry, or both.

\textbf{\texttt{documents}.} Defines the universe of acceptable document types. Each document has a unique \texttt{doc\_code} (e.g., "Type B" for Articles of Incorporation), a human-readable \texttt{name}, and a \texttt{description} used by the classification agent to identify document types. The \texttt{invalid\_if} field encodes invalidation rules as text (e.g., "Document is older than 12 months")—these are passed to the validation agent for policy checks. The \texttt{data\_source} enum indicates whether the document typically comes from external sources (client-provided), internal systems, or both.

\textbf{\texttt{data\_requirements}.} Defines the individual data fields that must be extracted from documents. Each field has a \texttt{data\_field} name in display format (e.g., "Date of Incorporation") and a \texttt{data\_schema} containing a JSON specification of the expected structure—including type, description, and nested structure for complex fields like addresses or authorized signatories. This schema is passed directly to the extraction agent, ensuring consistent output structure.

\textbf{\texttt{exemptions}.} Encodes conditional logic that modifies the journey checklist based on extracted data. Each exemption references a triggering \texttt{data\_requirement\_id} (the field whose value activates the exemption) and a target \texttt{doc\_requirement\_id} (the requirement being added or removed). The \texttt{is\_additive} boolean determines whether the exemption adds a new requirement or removes an existing one. The \texttt{description} field contains the condition logic in natural language, which the validation agent evaluates against extracted data.

\subsection{Join Tables}

Three join tables encode the many-to-many relationships that give the schema its flexibility:

\textbf{\texttt{acceptable\_documents}.} Links \texttt{doc\_requirements} to \texttt{documents}, specifying which document types can satisfy a given requirement. A single requirement (e.g., "Proof of Incorporation") may accept multiple document types (Articles of Incorporation, Certificate of Formation, Registration Extract), enabling the system to handle jurisdictional variations in document naming.

\textbf{\texttt{must\_contain}.} Links \texttt{data\_requirements} to \texttt{documents}, specifying which fields are inherently part of a document type regardless of the requirement context. For example, a Certificate of Incorporation must always contain an incorporation date and entity name. These fields are always extracted when the document type is identified.

\textbf{\texttt{doc\_data\_requirements}.} Links \texttt{data\_requirements} to \texttt{doc\_requirements}, specifying which fields must be extracted to satisfy a particular requirement. This allows the same document type to have different extraction schemas depending on which requirement it fulfills. For example, a Corporate Resolution used for "Authorized Signatories" requires extraction of signatory names and titles, while the same document type used for "Board Approval" may require extraction of resolution date and approval text.

\end{document}